\documentclass{ws-us}
\usepackage[sort,compress,super]{cite}
\usepackage{xcolor, soul}
\begin{document}
\setlength{\pdfpagewidth}{8.5in}
\setlength{\pdfpageheight}{11in}
\catchline{0}{0}{2013}{}{}

\markboth{Y.Liu et al.}{A Prompt-driven Task Planning Method for Multi-drones based on Large Language Model}

\title{A Prompt-driven Task Planning Method for Multi-drones \\based on Large Language Model}

\author{Yaohua Liu}

\address{ Guangdong Institute of Intelligence Science and Technology, \\ Zhuhai 519031, China\\
E-mail: liuyaohua@gdiist.cn}

\maketitle

\begin{abstract}
With the rapid development of drone technology, the application of multi-drones is becoming increasingly widespread in various fields. However, the task planning technology for multi-drones still faces challenges such as the complexity of remote operation and the convenience of human-machine interaction. To address these issues, this paper proposes a prompt-driven task planning method for multi-drones based on large language models. By introducing the Prompt technique, appropriate prompt information is provided for the multi-drone system. Leveraging the powerful semantic understanding capabilities of large language models, drones can accurately comprehend users' natural language task, enabling simpler, more efficient, and safer control of multi-drones, thereby enhancing the flight performance and application range of multi-drones. Test video: https://www.youtube.com/watch?v=yU1iviLBH24.
\end{abstract}

\keywords{multi-drone systems; large language model; prompt-driven control; deep learning.}
\begin{multicols}{2}
\section{Introduction}
Multi-drone systems involves complex tasks such as task description, task decomposition, task assignation and inter-drone coordination. The traditional task planning methods for multi-drone systems often rely on pre-defined algorithms, which may not be adaptable to dynamic and changing environments. As a result, there is a growing need for advanced task planning methods that can handle the complexity and variability of multi-drone systems. 

The first main challenges in task planning of multi-drone systems is the coordination and synchronization of multiple drones to achieve a common goal [\citen{pant2018fly}]. Each drone has its own capabilities, constraints, and sensing modalities, which need to be taken into account when designing control strategies. Moreover, the interactions between drones and their environment, as well as the interactions between drones themselves, further complicating the control problem. 

The second challenge lies in task allocation and path planning for multi-drone systems [\citen{shi2019multi}], [\citen{liu2023robust}]. Efficiently assigning tasks to drones and planning their trajectories while considering various constraints, such as energy consumption and collision avoidance, is a non-trivial problem. Traditional approaches often rely on centralized control or decentralized algorithms, which may not scale well with the increasing number of drones. 

To address these challenges, researchers have been exploring various task planning methods for multi-drone systems. These include decentralized control algorithms, swarm intelligence-based approaches, and machine learning techniques [\citen{guerber2021machine,schilling2019learning,heidari2023machine}]. However, there is still room for improvement in terms of efficiency, adaptability, and robustness [\citen{singh2023progprompt}],[\citen{tang2023saytap}]. 

Recently, there have been breakthroughs in natural language processing techniques, such as significant improvements in the ability of large language models (LLMs) to understand and generate text. As a result, incorporating large language models into task planning for robots has become a research topic aiming to enable top-level understanding of abstract and high-level natural language tasks of users [\citen{pramanick2020decomplex, venkatesh2021translating, yanaokura2022multimodal}]. 
However, the majority of the research has focused on the grasping capabilities of robotic arms [\citen{huang2022inner, zeng2022socratic, khan2023natural, kaynar2023remote}], with little attention given to task planning for drone swarms. Furthermore, these studies heavily rely on specific hardware and lack a satisfactory human-machine interaction experience [\citen{ding2023task, ding2022robot, liang2023code}]. Moreover, the reliance on specific datasets [\citen{ahn2022can, }] in most of these studies requires the collection of new data and retraining of models when attempting to transfer or extend them to different robotic environments.

In this context, the proposed prompt-driven control method based on large language models (LLMs) offers a promising approach. The use of large language models has gained significant attention in natural language processing tasks due to their ability to understand and generate human-like text. These models are pre-trained on vast amounts of text data, allowing them to capture rich semantic knowledge. By leveraging the semantic understanding capabilities of large language models, the control method can efficiently interpret user prompts and generate appropriate control instructions for multi-drone systems. This not only simplifies the control process but also allows for greater flexibility and adaptability in dynamic environments.

\section{Drones with LLM}
\begin{figure*}[ht]
\centerline{\includegraphics[width=4.7in]{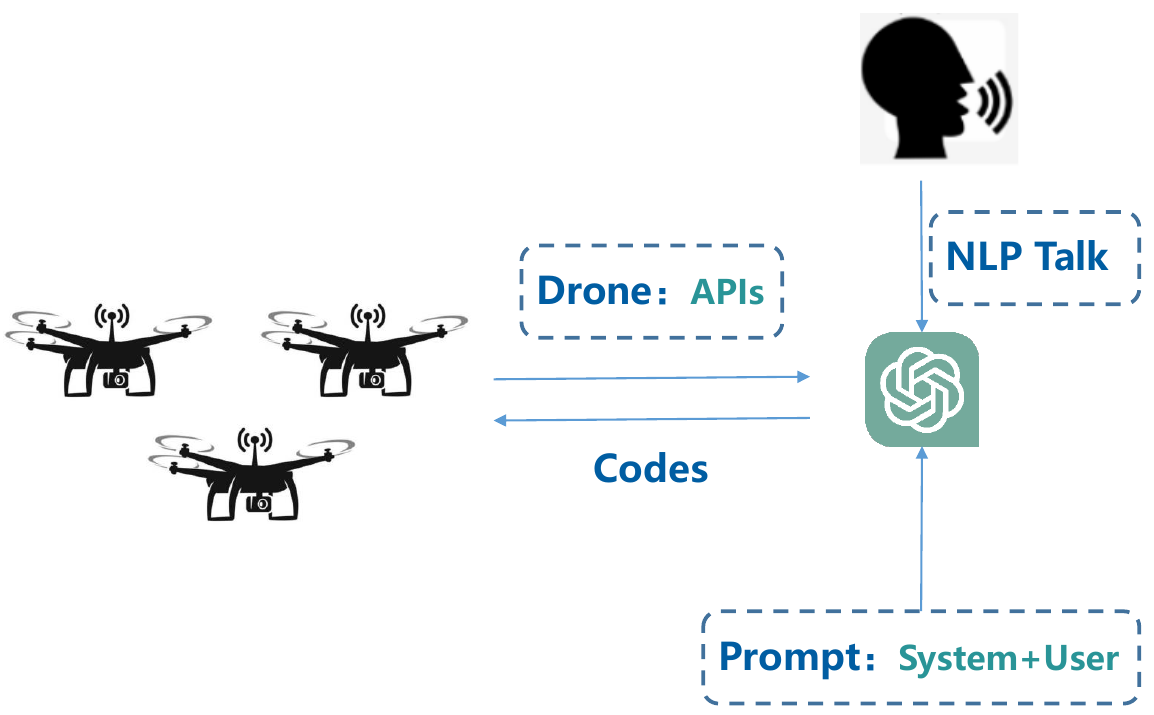}}
\caption{Framework of the prompt-driven control method for multi-drones based on LLM.}
\label{fig:pipeline}
\end{figure*}

Traditional approaches, such as collective intelligence-based methods or reinforcement learning-based methods[\citen{huttenrauch2019deep,xu2020reinforcement,hein2017particle}], often require substantial training data and computational resources. In contrast, our proposed method utilizes LLMs to achieve prompt-driven control and solves various drone-related tasks in a zero-shot method,  enabling more efficient control of multi-drone systems. As illustrated in Fig.\ref{fig:pipeline}, the pipeline of a prompt-driven control method for multi-drones based on LLM can be divided into the following four parts: constructing drone motion function library, designing system and user prompts, human-LLM interaction, executing the drone code generated by the LLM.

\subsection{Drone Motion Function Library}
A drone motion function library is a collection of pre-defined functions or algorithms that facilitate the control and movement of drones. The purpose of a drone motion function library is to provide a set of reusable and modular components that developers can utilize to build more complex drone control systems. These libraries often include functions for basic flight maneuvers like takeoff, landing, rotating and altitude control. They may also incorporate more advanced capabilities like path planning, formation flying, or dynamic obstacle avoidance.

The functions within a drone motion function library are typically implemented using mathematical models, control algorithms, and sensor data fusion techniques. They take into account various factors such as drone dynamics, environmental conditions, sensor inputs, and user commands to generate the desired drone motion.

By using a drone motion function library, developers can save time and efforts in implementing low-level control functionalities, as they can leverage pre-existing and tested functions. This also constrains the LLM to use the only the functions in the drone motion library so that the generated code can be recognized and executed by drones. 

\subsection{Prompt Design}

Writing well structured prompts is an essential part of ensuring accurate, high quality responses from a language model. In the context of a prompt-driven control system for multi-drone systems, there are two types of prompts: the system prompts and the user prompts.

\begin{figure*}[ht]
\centerline{\includegraphics[width=6.7in]{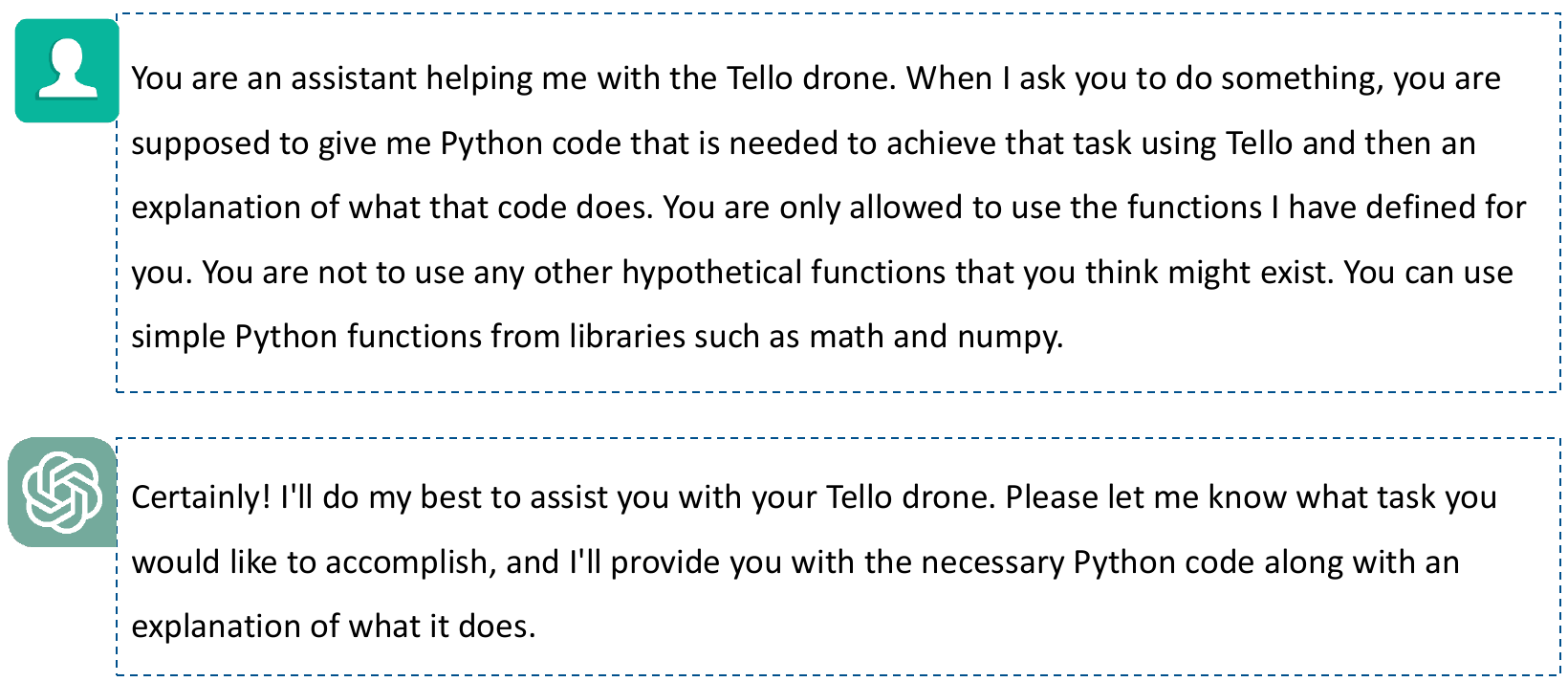}}
\caption{The system prompt.}
\label{fig:sp}
\end{figure*}

\begin{figure*}[ht]
\centerline{\includegraphics[width=6.7in]{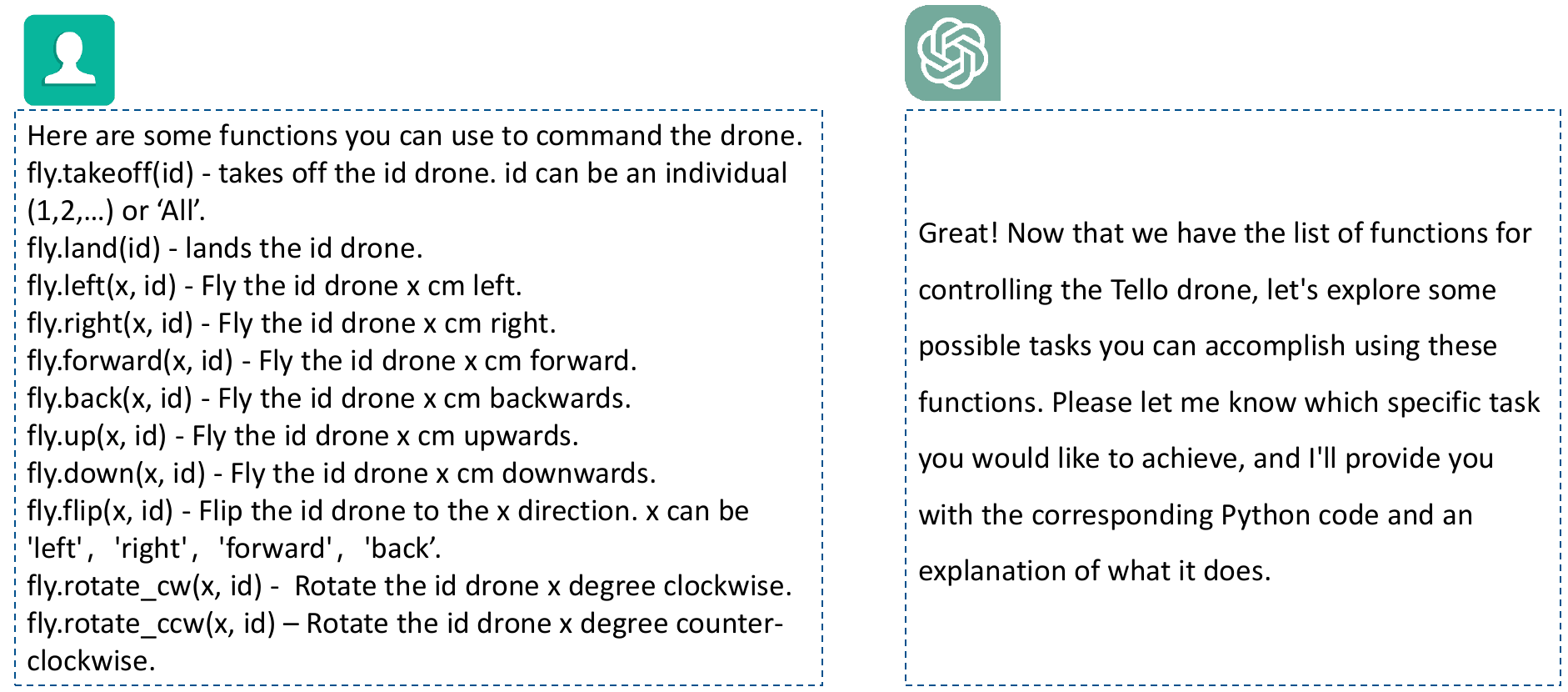}}
\caption{The user prompt.}
\label{fig:up}
\end{figure*}

The system prompts are predefined prompts or instructions that are designed to specify the function of the LLM. These prompts can change the role and function of the LLM, making it an expert in a certain field. As shown in Fig.\ref{fig:sp}, system prompts can include limitations such as "You are an assistant helping me with drones", "Use simple Python functions from libraries such as math and numpy", or "You are only allowed to use the functions I have defined for you". These prompts serve as guidelines for the LLM to follow and help in achieving specific objectives or tasks.
On the other hand, user prompts are prompts created by the user or operator of the multi-drone system. These prompts are usually in the form of natural language commands or requests provided by the user to the system. These prompts mainly present the executable functions for the drones followed by a brief explanation after each function. Taking a Tello drone as an example, Fig.\ref{fig:up} shows a typical user prompt. User prompts can include instructions like "Take off the No.1 drone", "Inspect a particular structure", or "Fly the No.1 and No.2 drones left simultaneously". The control system interprets these user prompts and generates appropriate system prompts to guide the drones accordingly.
The interaction between system prompts and user prompts forms the basis of the prompt-driven control approach. The system prompts provide the overall guidance and control for the multi-drone system, while the user prompts allow for flexible and intuitive communication between the user and the system. The control system analyzes and interprets the user prompts, generating system prompts that translate the user's intent into actionable instructions for the drones.

By incorporating both system prompts and user prompts, the prompt-driven control system enables effective and efficient control of multi-drone systems. It combines the advantages of automated control algorithms with the flexibility and adaptability of user input, allowing for dynamic and responsive drone operations in various scenarios.

\subsection{Human-LLM interaction}
Traditional command and control of multi-drone systems is implemented by ground control stations, where human operators design the flight mission by adding waypoints for the drones. This can be time consuming and does not apply to quickly developing scenarios where frequent task re-allocation of the multidorne systems are required. Thanks to the recent development of LLMs, Human-LLM interaction can be effectively applied to control and manage drone tasks. By leveraging the context understanding and code generation capabilities of LLMs, humans can communicate with drones in a more intuitive and natural way, enabling efficient and interactive task descriptions and drone formation control. 

Humans can provide high-level prompts or commands to the LLM to plan and coordinate drone tasks. During drone operations, humans can interact with the LLM in real-time to control the drones in a changing environment. If the operator needs to modify or adjust the ongoing drone tasks, they can provide new prompts to the LLM. There are other benefits in the human-LLM interaction. The LLM can assist humans in making informed decisions during drone operations. Humans can rely on LLMs to assess potential safety risks and mitigate them during drone operations. LLMs can assist in analyzing data collected by drones and generating reports. In summary, Human-LLM interaction enhances the control, coordination, and decision-making capabilities in drone operations. It enables more efficient and intelligent communication between humans and drones, leading to improved task execution, safety, and overall operational effectiveness.

\subsection{Drone Code Generation and Execution}
Drone code generation and execution by LLM involves utilizing the LLM's capabilities to generate code for drone tasks and then executing that code on the drones. The LLM processes the provided prompts and generates code based on the desired drone tasks. The generated code can include flight control commands, sensor data processing, image or video capture, navigation instructions, or any other specific actions required for the tasks. Next, the code needs to be executed on the drones. 

During code execution, it is essential to establish communication channels between the operator, the LLM, and the drones. This allows for real-time feedback and status updates. The drones can send telemetry data, sensor readings, or images back to the operator or the LLM, enabling monitoring and decision-making based on the drone's current state. It is crucial to implement error-handling mechanisms and safety protocols when performing code generated by LLM. This includes checking for potential errors or exceptions in the code, implementing fail-safe mechanisms, and ensuring compliance with safety regulations and guidelines. Throughout the execution of the generated code, the operator or the LLM should monitor the drones' behavior and ensure that they are performing the desired tasks correctly. This may involve adjusting parameters, providing additional instructions, or modifying the code if necessary. Once the code execution is complete, the operator or the LLM can analyze the collected data, images, or other outputs from the drones. This analysis can help evaluate the success of the tasks, identify any issues or anomalies, and provide insights for future improvements.

Performing the drone code generated by LLM requires a careful and systematic approach to ensure the safe and accurate execution of the intended tasks. It is crucial to follow best practices, maintain situational awareness, and adhere to applicable regulations to ensure the successful operation of drones.

\section{Experiments}
The Tello drone, developed by Ryze Tech in collaboration with DJI, is a small and lightweight drone designed for recreational and educational purposes. It is often used for beginners or individuals interested in learning about drones and programming and offers various features and can be controlled using a smartphone or compatible programming languages. Thus, we chose the Tello drone as our experimental subject and conducted experiments on LLM control of a single drone, synchronized control of multiple drones, and asynchronous control of multiple drones.

\subsection{LLM control of a single drone}
\begin{figure*}[ht]
\centerline{\includegraphics[width=6.7in]{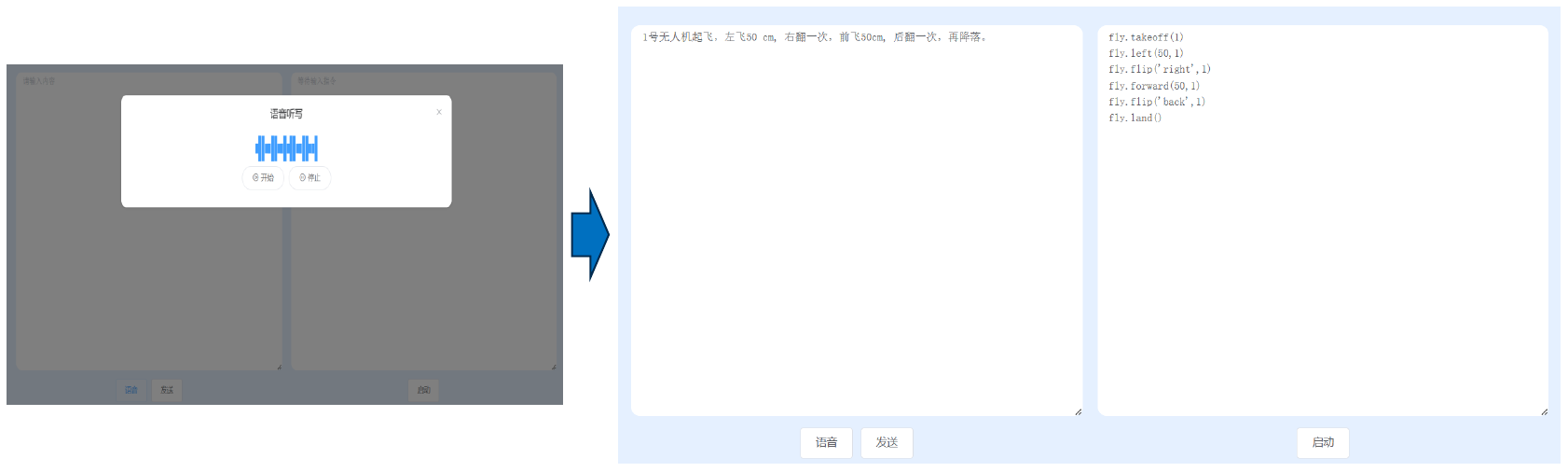}}
\caption{The voice-text user interface.}
\label{fig:gui}
\end{figure*}

\begin{figure*}[ht]
\centerline{\includegraphics[width=6.7in]{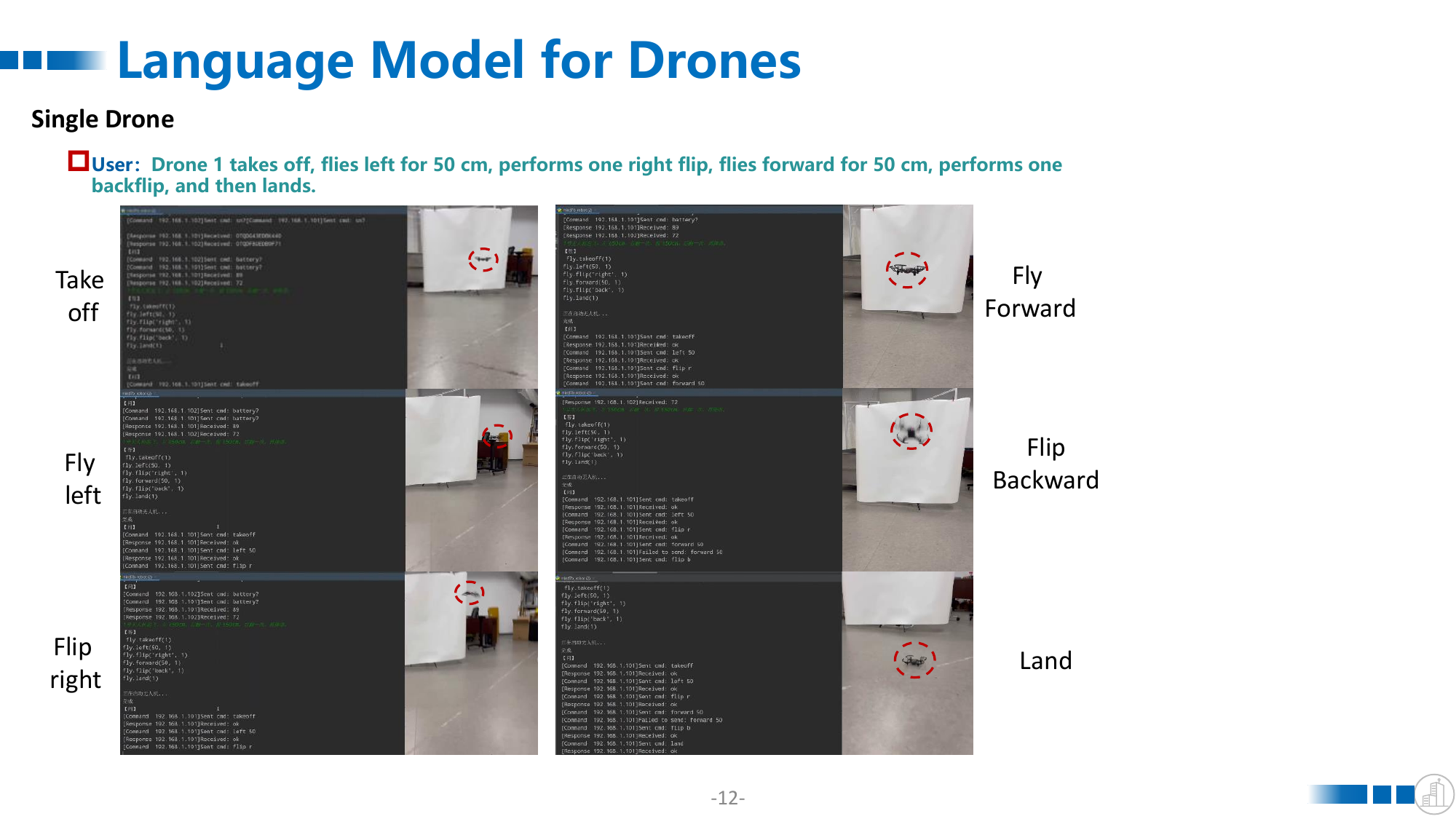}}
\caption{The single drone tasks.}
\label{fig:user1}
\end{figure*}

Firstly, we set the Tello drone to AP mode and connected the drone and the laptop to the same router via WiFi, creating a local area network (LAN). To enhance user interaction, we developed a voice interaction GUI interface. As shown in Fig.\ref{fig:gui}, users can engage in dialogue and communication with the LLM by clicking the voice button. The underlying voice function library converts speech into text messages and sends them to the LLM. Leveraging its powerful natural language understanding capabilities, the LLM analyzes and extracts the user's task intentions, combines them with prompts, and generates corresponding drone action codes. Finally, the LLM sends the generated code to the Tello drone via the UDP network protocol for execution, successfully completing the user's advanced tasks.
To conduct the test, we verbally expressed the single drone task shown in Fig.\ref{fig:user1} in Chinese. Theoretically, it can be done in any official language internationally, such as English, French, and so on.
From the experimental results, it can be observed that after understanding the user's voice task, the LLM first generates the takeoff command function, then generates the drone's control command to fly 50cm to the left. It then performs a right flip, generates the code for flying forward 50cm, followed by a backward flip, and finally performs a landing. Actual flight tests have also verified that controlling drones through LLM can effectively accomplish advanced tasks as desired by the user.

\subsection{Synchronized control of multiple drones}
\begin{figure*}[ht]
\centerline{\includegraphics[width=6.7in]{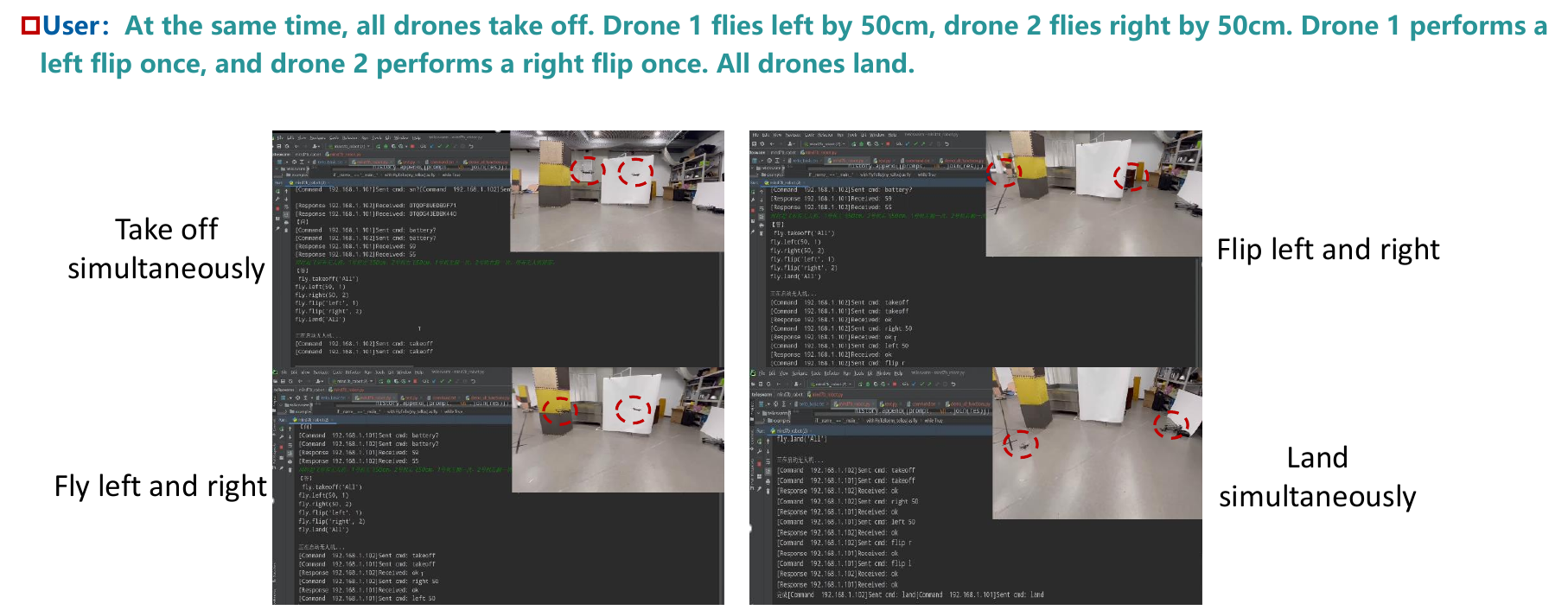}}
\caption{The Multi-Drones: Synchronous Scenario.}
\label{fig:user2}
\end{figure*}

To further test the effectiveness of LLM in controlling multiple Tello drones, we conducted separate experiments on both synchronous and asynchronous control of two drones. The experimental procedure is similar to that of a single drone experiment, with the difference being that we assigned a unique ID to each drone through an IP address, making it convenient for LLM to control each drone separately. As shown in Fig.\ref{fig:user2}, we verbally expressed specific tasks in Chinese, and after LLM received our task commands, it generated the code to simultaneously control the drones for takeoff, left-right flight and flips. Each action command code, generated by LLM after understanding the user's task, is correct and executed well by the Tello drones.
\subsection{Asynchronous control of multiple drones}
\begin{figure*}[ht]
\centerline{\includegraphics[width=6.7in]{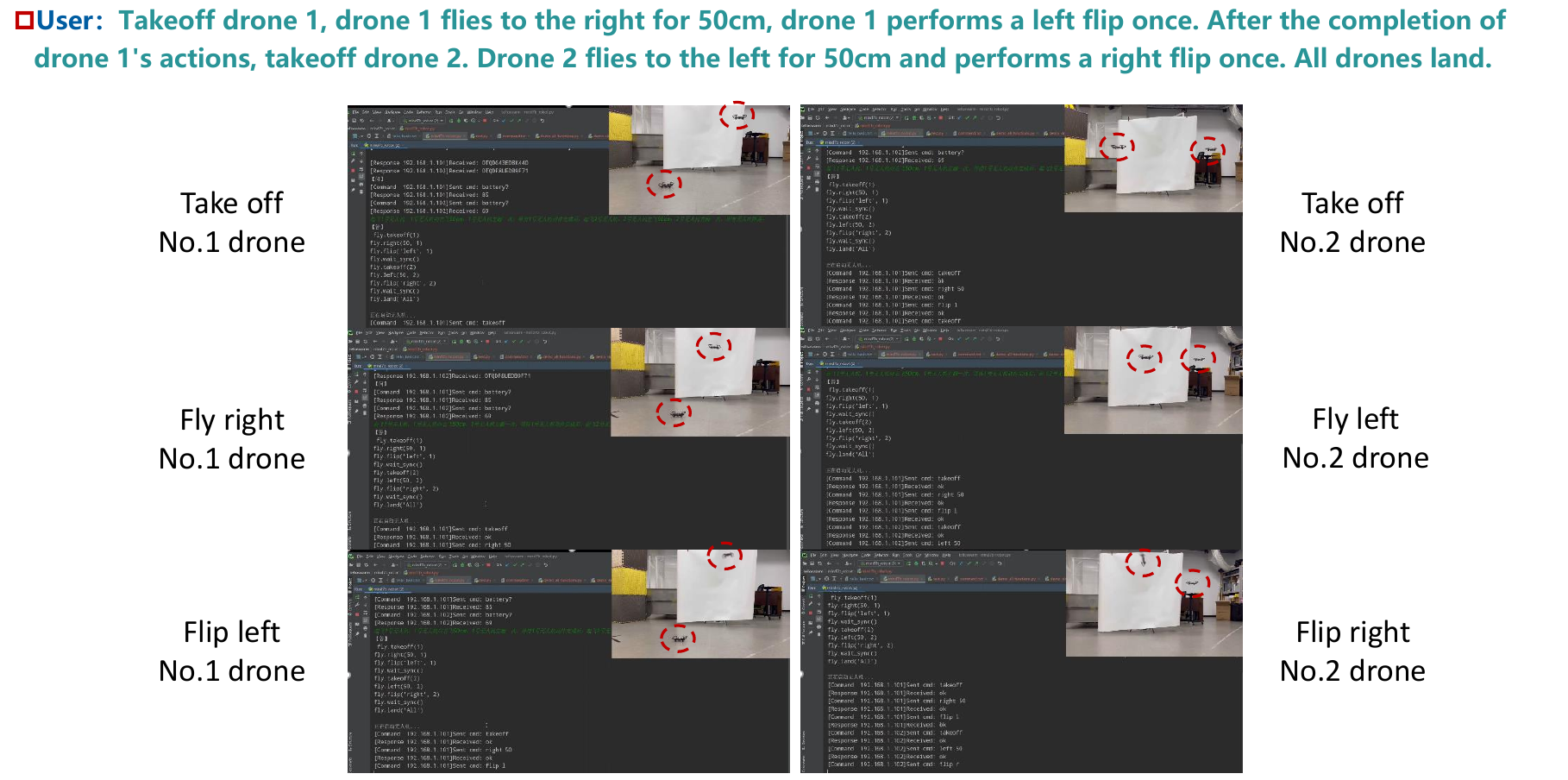}}
\caption{The Multi-Drones: Asynchronous Scenario.}
\label{fig:user3}
\end{figure*}

The purpose of designing the asynchronous control experiment is mainly to test LLM's ability to handle time sequences. As shown in Fig.\ref{fig:user3}, unlike the synchronous control experiment, the drone action sequence in the asynchronous control experiment has a temporal order. The user communicates with LLM in Chinese, saying "Take off drone 1, drone 1 flies to the right 50cm, drone 1 performs a left flip. After the completion of drone 1's actions, take off drone 2, drone 2 flies to the left 50cm, drone 2 performs a right flip, and all drones land." In this dialogue, the user's tasks have a clear time sequence, such as drone 2 needing to perform its actions after drone 1 completes its actions. From the experimental results, it can be seen that even for complex time sequence tasks, LLM can still understand them well and generate correct codes for successful execution by the Tello drones.
\section{Conclusion}
This paper proposes a user-friendly and intuitive way to control multiple drones. It eliminates the need for complex manual control interfaces and allows operators to communicate with drones in a more natural and efficient manner. The experiments show that the LLM can accurately interpret user commands and generate appropriate control codes for multiple drones to execute coordinated actions. It showcases the potential of large language models in enhancing human-drone interaction and facilitating the deployment of drone systems in various applications.

\bibliographystyle{ieeetr}

\bibliography{ref}



\end{multicols}
\end{document}